\documentclass[
]{ceurart}

\usepackage{listings}
\usepackage{svg}
\usepackage{subcaption}
\usepackage{hyperref}
\usepackage{tcolorbox}
\usepackage{arydshln}
\usepackage{makecell}
    \usepackage{tikz}
    
    \usetikzlibrary{positioning, arrows.meta, fit, calc,
                    shapes.geometric, shadows.blur, backgrounds}
    \usepackage{amssymb}
\definecolor{priceLight}{HTML}{E4D9EE}
\definecolor{priceBold}{HTML}{7B5BA8}
\definecolor{newsLight}{HTML}{D1E1F2}
\definecolor{newsBold}{HTML}{3F76A8}
\definecolor{stateLight}{HTML}{E3F0DA}
\definecolor{stateBold}{HTML}{5E9A34}
\definecolor{envLight}{HTML}{FCE6C7}
\definecolor{envBold}{HTML}{CF8427}
\definecolor{agentLight}{HTML}{F9D2D2}
\definecolor{agentBold}{HTML}{B94B4B}
\definecolor{zoomRed}{HTML}{C24A4A}
\definecolor{frozenBlue}{HTML}{5A9BD4}
\definecolor{fireOrange}{HTML}{E75D1F}

\newcommand{\frozenicon}{%
    \tikz[baseline=-0.4ex,x=1mm,y=1mm]{%
        \foreach \a in {0,60,120}{\draw[frozenBlue, line width=0.55pt] (\a:0.9) -- (\a+180:0.9);}
        \foreach \a in {0,60,120,180,240,300}{%
            \draw[frozenBlue, line width=0.55pt] (\a:0.45) -- ($(\a:0.45)+(\a+60:0.28)$);%
            \draw[frozenBlue, line width=0.55pt] (\a:0.45) -- ($(\a:0.45)+(\a-60:0.28)$);%
        }%
    }%
}

\newcommand{\trainableicon}{%
    \tikz[baseline=-0.4ex,x=1mm,y=1mm]{%
        \fill[fireOrange]
            (0,0)
            .. controls (0.6,0.3) and (0.9,0.9) .. (0.4,1.5)
            .. controls (0.55,1.0) and (0.0,1.0) .. (-0.15,1.7)
            .. controls (-0.7,1.2) and (-0.6,0.4) .. (-0.3,0.1)
            .. controls (-0.15,-0.1) and (0.1,-0.1) .. (0,0);
        \fill[yellow!90!orange]
            (-0.05,0.3)
            .. controls (0.3,0.5) and (0.35,1.0) .. (0.05,1.3)
            .. controls (0.15,0.9) and (-0.2,0.9) .. (-0.25,0.5)
            .. controls (-0.25,0.3) and (-0.1,0.25) .. (-0.05,0.3);
    }%
}

\begin{document}

\copyrightyear{2026}
\copyrightclause{Copyright for this paper by its authors.
  Use permitted under Creative Commons License Attribution 4.0
  International (CC BY 4.0).}

\conference{CLEF 2026 Working Notes, 21 – 24 September 2026, Jena, Germany}

\title{CLaC @ FinMMEval 2026 Task 3: Sentiment-Augmented Deep Reinforcement Learning for Active Trading -- An Alpha-Reward Approach}

\author[]{Andrei Neagu}[%
email=andrei.neagu@concordia.ca,
]
\cormark[1]
\address[]{Department of Computer Science and Software Engineering, Concordia University, Montréal, Canada}

\author[]{Eeham Khan}[%
email=eeham.khan@concordia.ca,
]

\author[]{Leila Kosseim}[%
email=leila.kosseim@concordia.ca,
]

\cortext[1]{Corresponding author: andrei.neagu@concordia.ca}

\begin{abstract}
    This paper presents our system for Task 3 of the CLEF 2026 FinMMEval Lab, which requires participants to submit a daily trading decision (\textsc{long}, \textsc{flat}, or \textsc{short}) for Bitcoin (BTC) and Tesla (TSLA) based on news articles and historical market pricing data. We frame the problem as a discrete-action Markov Decision Process and train four Deep Reinforcement Learning (DRL) algorithms: Policy Gradient (PG), Proximal Policy Optimization (PPO), Deep $Q$-learning (DQL), and Deep Deterministic Policy Gradient (DDPG), using a rich feature set that combines technical indicators (EMA, RSI, MACD, BollingerB, volume change), cyclical date encodings, and daily sentiment scores derived from news articles using LLaMA 3.2 1B. To reduce overfitting and align the training objective with the goal of outperforming a buy-and-hold baseline, we introduce an \emph{alpha reward} that replaces the raw log-return with excess return over the market, and we randomize episode start days during training. Hyperparameters are optimized over 180 trials per algorithm-asset pair using Ray Tune, with model selection based on validation Sharpe ratio (SR) and early stopping. Evaluation on the CLEF Task 3 test set demonstrates that DDPG yields the best performance across both assets. Note, however, that DQL was selected a priori for the live competition endpoint based on its highest validation Sharpe ratio; the endpoint model was deliberately selected blind to the test period to avoid selection bias. For TSLA, DDPG and DQL achieved cumulative returns of 54.96\% and 52.62\% respectively, substantially beating the 16.45\% buy-and-hold baseline. On the more challenging BTC test set, DDPG mitigated severe market losses, achieving a positive return of 1.58\% against a baseline decline of -34.27\%. Results reveal a pronounced validation-to-test generalization gap, pointing to the difficulty of adapting policies trained in a bull market validation period to bear market test conditions.
\end{abstract}


\maketitle

\section{Introduction}
\label{se:introduction}
Financial markets are among the most complex sequential decision-making environments, combining noisy price dynamics and information sources such as global and asset specific news and economic indicators. Traditional quantitative strategies rely on hand-crafted rules or statistical models that struggle to capture non-linear interactions in multi-modal settings. Deep Reinforcement Learning (DRL) methods offer an alternative better suited to learning an optimal trading strategy adapted to such interactions. In DRL, an agent learns a policy that maps market observations directly to trading actions by maximizing a cumulative reward signal \cite{MnihDQN}. 

Task 3 of the CLEF~2026 FinMMEval Lab \citep{Xie2026clef2026finmmeval, FinMMEvalTask3Overview2026, qian2025whenagentstrade} aims at exploring dynamic market environments through real-time evaluation. Participants provide a web endpoint that receives daily market context consisting of prices, news articles, momentum, and optional SEC filings, and must respond with one of three actions: \textsc{long}, \textsc{flat}, or \textsc{short}. Positions are updated daily and performance is assessed by cumulative return (CR), Sharpe ratio (SR), maximum drawdown (MDD), and volatility on two assets: Bitcoin (BTC) and Tesla (TSLA).

Our work tackles this challenge using four DRL algorithms trained and validated on historical data spanning from 2010 to 2024. We make three principal contributions:

\begin{enumerate}
    \item We demonstrate that a multi-modal feature representation combining price-derived technical indicators, LLM-based daily news sentiment scores, and cyclical calendar encodings provides an effective state signal for DRL-based trading agents.
    \item We introduce and analyze an alpha reward formulation that explicitly trains agents to outperform the \textit{buy-and-hold} baseline rather than to maximize absolute portfolio value, and show that it (i) equals log-alpha when summed over an episode, (ii) shares optimal policies with the raw log-return reward under exogenous market dynamics, and (iii) acts as a variance-reducing control variate on the reward signal. Combined with random episode starts, our alpha reward reduces overfitting and aligns the training objective with the evaluation criterion.
    \item We provide a comprehensive comparison of four DRL algorithms (PG, PPO, DQL, and DDPG) with hyperparameter optimization and Sharpe-based early stopping, and show that DDPG achieves the most robust performance across both assets while validation-based model selection remains sensitive to regime shifts between bull and bear markets.\footnote{All code, trained model checkpoints, hyperparameter configurations, and data preprocessing scripts (including the regex filters, deduplication, and sentiment-scoring pipeline) will be released upon paper acceptance to ensure full reproducibility.}
\end{enumerate}

\section{Related Work}
\label{se:relatedWork}
Early work applying Reinforcement Learning (RL) to market trading dates back to \citet{Moody1998DRL_Trading}, who proposed the use of recurrent neural networks trained on cumulative return. \citet{MnihDQN} later showed that Deep $Q$-Networks (DQN) could match or surpass human performance in Atari games, inspiring a wave of applications to financial markets. \citet{Jiang2017Portfolio} formulated cryptocurrency portfolio management as a continuous-action Deep Reinforcement Learning (DRL) problem, showing that an LSTM-based deterministic Policy Gradient (PG) agent could outperform traditional statistical rebalancing strategies. 

Actor-critic methods have also seen growing adoption in trading. Building on foundational actor-critic architectures, \citet{pmlr-v48-mniha16} introduced Asynchronous Advantage Actor-Critic (A3C), which stabilizes learning and improves sample efficiency by deploying multiple parallel workers to interact with their own environments asynchronously. To simplify this architecture, a synchronous variant, Advantage Actor-Critic (A2C), was then developed to coordinate parallel workers and update the global network synchronously while maintaining comparable performance. Additionally, \citet{Schulman2017PPO} introduced Proximal Policy Optimization (PPO), a stable on-policy method widely adopted for financial applications due to its clipped surrogate objective. \citet{Lillicrap2016continuous} proposed Deep Deterministic Policy Gradient (DDPG), which extends Deep $Q$-Learning (DQL) to continuous action spaces via a deterministic policy with soft target updates. \citet{Liu2018DRLPortOpt} applied DDPG to stock portfolio management, and demonstrated that it outperformed the Dow Jones buy-and-hold strategy in backtesting.


Following these results, \citet{Yang2020EnsembleTrading} employed an ensemble approach, combining PPO, DDPG, and A3C, to trade a portfolio comprised of the 30 constituent stocks of the Dow Jones Industrial Average. The authors constantly evaluate each individual algorithm, switching to the best performing one after $N$ steps. They showed that, while their ensemble method produces a higher Sharpe ratio (SR), PPO achieves a higher cumulative return. \citet{Ye2024SentimentBasedEnsemble} have also employed an ensemble method using the DDPG, A2C and PPO algorithms, but only switch to the best performing algorithm when sentiment scores exceed a certain threshold. Their results show that this selection method outperforms the $N$-step selection method.
More recently, \citet{Du2024DRL_CNN_Trading} employed DQL with sentiment analysis for trading a portfolio comprised of Chinese stocks, finding that their method outperforms both standard DQL and traditional statistical methods.

The integration of natural language information into trading agents has gained significant interest due to the increasing availability of pre-trained and large language models (LLMs). \citet{Araci2019FinBERT} proposed FinBERT, a BERT model fine-tuned on financial corpora that produces sentiment scores that can be used as features in prediction models. 
Recent research has also explored the potential for LLMs to act as financial decision-makers \cite{Yu2023LLM_Prediction,LopezLira2025ChatGPT_Forecast}.
Our work follows this line of research, as we use an open-source LLM for article-level sentiment scoring combined with DRL algorithms for action selection. 

\section{Background}
\label{se:background}

We model the trading problem as a Markov Decision Process (MDP) defined by the tuple $(\mathcal{S},\mathcal{A},\mathcal{R},\mathcal{P},\gamma)$ \cite{Sutton1998}, where:
\begin{itemize}
    \item $\mathcal{S}$ is the state space: a vector of market features observed at each trading day.
    \item $\mathcal{A} = \{\textsc{short}\ (-1),\ \textsc{flat}\ (0),\ \textsc{long}\ (1)\}$ is the discrete action space.
    \item $\mathcal{R}(s,a,s')$ is the reward function associated with taking action $a \in \mathcal{A}$ in state $s \in \mathcal{S}$ and transitioning to the next state $s' \in \mathcal{S}$.
    \item $\mathcal{P}(s'|s,a)$ is the (unknown) market transition probability, representing the stochastic evolution of the financial environment. We assume the agent's actions do not influence future market states (price-taker).
    \item $\gamma \in [0,1)$ is the discount factor, determining the present value of future rewards and ensuring that the expected return $G_t$ remains finite.
\end{itemize}

At each time step $t \in [1, T]$, the agent observes state $s_t$, selects action $a_t \sim \pi_\theta(s_t)$, transitions to $s_{t+1}$, and receives a reward $r_t$. The objective is to find the policy $\pi_\theta$ that maximizes the expected discounted return $G_t=\sum_{k=0}^{T-t}\gamma^k r_{t+k}$.

To translate these discrete actions into financial performance, we evaluate the agent based on a portfolio that evolves multiplicatively. At each time step $t$, the agent's previously chosen action dictates its current market position $p_{t} \in \{-1,0,1\}$ (\textsc{long}, \textsc{flat}, or \textsc{short}). As the market transitions, the agent observes the daily price return $return_t=(c_t-c_{t-1})/c_{t-1}$, where $c_t$ is the closing price on day $t$. Consequently, the portfolio value $V_t$ updates as follows:
\begin{equation}
    V_t=V_{t-1} \cdot (1+p_{t-1} \cdot return_t) \cdot \big(1-\delta \cdot\mathds{1}_{\{a_t \neq p_{t-1}\}}\big)
\end{equation}
where $\delta=0.002$ is the transaction cost, and $\mathds{1}_{\{a_t \neq p_{t-1}\}}$ is an indicator function that equals $1$ when the agent's new action $a_t$ differs from its previous position $p_{t-1}$, i.e. when a trade is executed, and 0 otherwise, so that the cost is incurred only on position changes.

\section{Methodology}
\label{se:methodology}

\begin{figure}[t]
    \centering
    \resizebox{\linewidth}{!}{%
    \begin{tikzpicture}[
        font=\sffamily\scriptsize,
        >={Latex[length=1.6mm, width=1.3mm]},
        box/.style={rectangle, rounded corners=2.5pt, line width=0.45pt, align=center, inner sep=2.5pt,
                    blur shadow={shadow blur steps=4, shadow blur extra rounding=0.5pt,
                                 shadow opacity=18, shadow xshift=0.3mm, shadow yshift=-0.3mm}},
        priceBox/.style={box, fill=priceLight, draw=priceBold},
        newsBox/.style={box, fill=newsLight, draw=newsBold},
        stateBox/.style={box, fill=stateLight, draw=stateBold},
        envBox/.style={box, fill=envLight, draw=envBold},
        agentBox/.style={box, fill=agentLight, draw=agentBold},
        algoChip/.style={rectangle, rounded corners=1.8pt, fill=white, draw=agentBold,
                         line width=0.4pt, minimum width=13mm, minimum height=7mm,
                         inner sep=1pt, font=\sffamily\scriptsize\bfseries, align=center},
        featCell/.style={rectangle, rounded corners=1pt, fill=white, draw=stateBold!80,
                         line width=0.3pt, minimum height=4mm, minimum width=9mm,
                         inner sep=1pt, font=\sffamily\tiny},
        tokenP/.style={rectangle, rounded corners=0.5pt, fill=priceLight, draw=priceBold,
                       line width=0.25pt, minimum size=3mm, inner sep=0},
        tokenN/.style={rectangle, rounded corners=0.5pt, fill=newsLight, draw=newsBold,
                       line width=0.25pt, minimum size=3mm, inner sep=0},
        flow/.style={->, black!55, line width=0.55pt, shorten <=0.3mm, shorten >=0.3mm},
        zoomConn/.style={draw=zoomRed, dash pattern=on 1.5pt off 1pt, line width=0.55pt},
        legendBox/.style={rectangle, rounded corners=2pt, draw=black!35, fill=white,
                          dash pattern=on 1pt off 1pt, line width=0.35pt, inner sep=2.5pt,
                          font=\sffamily\tiny}
    ]

    \foreach \i in {0,...,5}{\node[tokenP] at ({\i*3.6mm},0) {};}
    \node[font=\sffamily\tiny\bfseries, priceBold] at (0.9,-0.55) {Price \& Volume};
    \foreach \i in {0,...,5}{\node[tokenN] at ({3.3+\i*3.6mm},0) {};}
    \node[font=\sffamily\tiny\bfseries, newsBold] at (4.2, 0.02) {News Article};

    \node[priceBox, minimum width=26mm, minimum height=11mm]
        (tech) at (0.9, 1.65)
        {\textbf{Technical}\\\textbf{Indicators}\\[-0.3mm]{\tiny EMA, RSI, MACD,}\\[-0.3mm]{\tiny Bollinger, Vol\,$\Delta$}};
    \node[newsBox, minimum width=26mm, minimum height=11mm]
        (sent) at (4.2, 1.65)
        {\textbf{Sentiment Analysis}~\frozenicon\\[-0.3mm]{\tiny log-prob extraction}\\[-0.3mm]{\tiny $s=P(+)-P(-)$}};
    \node[box, fill=gray!8, draw=gray!55, minimum width=18mm, minimum height=11mm]
        (cal) at (7.85, 1.65)
        {\textbf{Calendar}\\[-0.3mm]{\tiny DoW, DoM, MoY}\\[-0.3mm]{\tiny (sin/cos)}};

    \draw[flow] (0.9,0.2) -- (tech.south);
    \draw[flow] (4.2,0.2) -- (sent.south);

    \node[stateBox, minimum width=105mm, minimum height=13mm]
        (state) at (2.55, 3.7) {};
    \node[font=\sffamily\bfseries, anchor=north]
        at ($(state.north)+(0,-0.5mm)$) {State Vector $s_t \in \mathbb{R}^{15}$};
    \node[featCell, anchor=west] (c1) at ($(state.west)+(2mm,-2mm)$) {Price \& Sentiment};
    \node[featCell, right=2mm of c1] (c2) {Momentum \& Trend (EMA, RSI, MACD)};
    \node[featCell, right=2mm of c2] (c3) {Volatility \& Volume};
    \node[featCell, right=2mm of c3] (c4) {Time Cycles \& Pos. $p_{t-1}$};

    \draw[flow] (tech.north) -- ++(0,2mm) -| ($(state.south)+(-30mm,0)$);
    \draw[flow] (sent.north) -- ++(0,2mm) -| ($(state.south)+(0,0)$);
    \draw[flow] (cal.north)  -- ++(0,2mm) -| ($(state.south)+(28mm,0)$);

    \node[envBox, minimum width=80mm, minimum height=13mm]
        (env) at (2.55, 5.45)
        {\textbf{Trading Environment}\\[0.2mm]{\scriptsize Actions $\{-1,0,+1\}$\quad $\delta=0.2\%$ cost (train)\quad random start (train)}\\[0.2mm]{\scriptsize $\displaystyle \alpha$-reward:\; $r_t = \log(V_t/V_{t-1}) - \log(c_t/c_{t-1})$}};
    \draw[flow] (state.north) -- (env.south);

    \node[agentBox, minimum width=80mm, minimum height=14mm]
        (agent) at (2.55, 7.35) {};
    \node[font=\sffamily\scriptsize\bfseries, anchor=north]
        at ($(agent.north)+(0,-0.6mm)$) {DRL Agent~\trainableicon};
    \node[algoChip, anchor=west] (a1) at ($(agent.west)+(4mm,-1.6mm)$) {PG};
    \node[algoChip, right=6mm of a1] (a2) {PPO};
    \node[algoChip, right=6mm of a2] (a3) {DQL};
    \node[algoChip, right=6mm of a3] (a4) {DDPG};
    \draw[flow] (env.north) -- (agent.south);

    \node[font=\sffamily\footnotesize\bfseries] (action) at (2.55, 8.7)
        {Action $a_t \in \{\textsc{long}, \textsc{flat}, \textsc{short}\}$};
    \draw[flow] (agent.north) -- (action.south);

    \begin{scope}[xshift=112mm]
    \node[rectangle, rounded corners=3pt, draw=zoomRed, line width=0.6pt,
          dash pattern=on 2pt off 1.2pt, fill=white,
          minimum width=50mm, minimum height=86mm, anchor=south]
        (zoomBG) at (2.5, -0.6) {};
    \node[font=\sffamily\scriptsize\bfseries, anchor=north]
        at ($(zoomBG.north)+(0,-1.5mm)$) {LLaMA 3.2 Sentiment Scoring};

    \node[newsBox, minimum width=40mm, minimum height=7mm]
        (zArt) at (2.5, 0.5) {\textbf{Article} (truncated to 300 words)};
    \node[box, fill=newsLight!40, draw=newsBold!60, minimum width=40mm, minimum height=11mm,
          font=\sffamily\tiny, above=4mm of zArt, align=left]
        (zPrompt) {\textbf{Structured prompt}\\[0.3mm]\texttt{[System]} classify for \{asset\}:\\\texttt{\ \ }positive / negative / neutral\\\texttt{[User]} \{article\}};
    \draw[flow] (zArt) -- (zPrompt);

    \node[box, fill=gray!10, draw=gray!55, minimum width=40mm, minimum height=16mm,
          above=4mm of zPrompt]
        (zLlama) {\textbf{LLaMA 3.2 1B Instruct}~\frozenicon\\[0.5mm]{\scriptsize single forward pass}\\{\scriptsize no generation}};
    \draw[flow] (zPrompt) -- (zLlama);

    \node[box, fill=priceLight!50, draw=priceBold!70, minimum width=40mm, minimum height=8mm,
          above=4mm of zLlama, font=\sffamily\tiny]
        (zLogits) {\textbf{Last-token logits}~$\in \mathbb{R}^{|V|}$};
    \draw[flow] (zLlama) -- (zLogits);

    \node[box, fill=envLight!80, draw=envBold!80, minimum width=40mm, minimum height=10mm,
          above=4mm of zLogits, font=\sffamily\tiny]
        (zSoft) {\textbf{Softmax over token IDs}\\\{\texttt{positive},\ \texttt{negative},\ \texttt{neutral}\}};
    \draw[flow] (zLogits) -- (zSoft);

    \node[box, fill=stateLight, draw=stateBold, minimum width=40mm, minimum height=8mm,
          above=4mm of zSoft]
        (zScore) {\textbf{Sentiment score}\\[0.2mm]$s = P(+) - P(-) \in [-1,1]$};
    \draw[flow] (zSoft) -- (zScore);
    \end{scope}

    \draw[zoomConn] (sent.north east) -- (zoomBG.north west);
    \draw[zoomConn] (sent.south east) -- (zoomBG.south west);

    \node[legendBox] (legend) at (8.75, 7.15) {\trainableicon\ Trainable\quad\frozenicon\ Frozen};

    \end{tikzpicture}%
    }
    \caption{System overview. \textbf{Left:} three parallel processing streams, technical indicators derived from price and volume (purple) and daily news sentiment from a frozen LLaMA-3.2-1B Instruct model (blue), together with cyclical calendar encodings, compose a $15$-dimensional state vector $s_t$. The state is fed as input by a custom trading environment whose $\alpha$-reward incentivises outperforming the buy-and-hold baseline, and by one of four DRL agents (PG, PPO, DQL, DDPG) that outputs a daily action. \textbf{Right:} Detail of the sentiment scoring module, which performs a single forward pass of the LLaMA backbone model, extracts the logits at the final token position, and applies a softmax restricted to the token IDs for \{positive, negative, neutral\}, yielding a bounded score $s \in [-1,1]$.}
    \label{fig:system}
\end{figure}
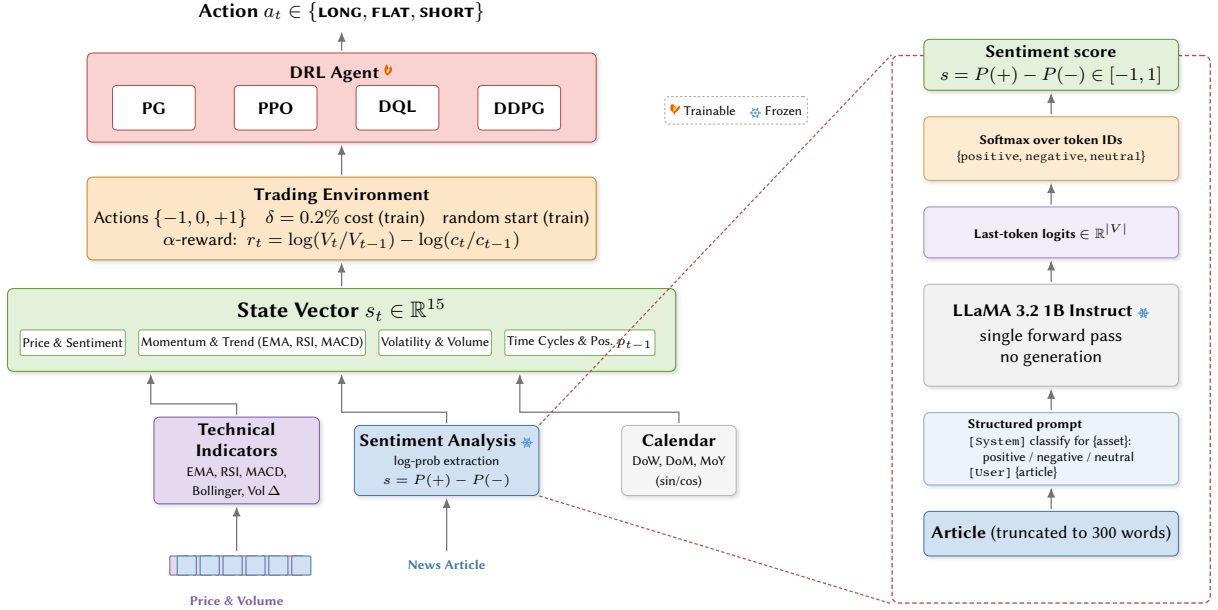

Figure \ref{fig:system} illustrates the overall architecture of our proposed trading framework. The system processes historical pricing and textual news data (see Section \ref{se:data}), extracts daily sentiment using a zero-shot LLM (see Section \ref{se:sentiment}), and constructs a 15-dimensional state feature vector (see Section \ref{se:features}). This state representation informs a custom Markov Decision Process trading environment, where one of four DRL agents is trained to maximize a market-relative alpha reward (see Section \ref{se:DRL}).





\subsection{Data}
\label{se:data}

\paragraph{External Data.}
In order to supplement the provided training data, we augment it with externally sourced data. Daily closing prices and trading volumes are downloaded via the Yahoo Finance (\texttt{yfinance}\footnote{\href{https://pypi.org/project/yfinance/}{https://pypi.org/project/yfinance/}}) library. For TSLA, the data spans 2,835 trading days from 2010-06-29 to 2022-12-30, covering the asset's entire history up to the validation cutoff. For BTC-USD, the data covers 2,664 trading days from 2014-09-17 (the earliest date available on Yahoo Finance) to 2022-12-30. We align each trading day with news articles from Brian Ferrell's Financial News Multi-Source Hugging Face dataset\footnote{\href{https://huggingface.co/datasets/Brianferrell787/financial-news-multisource}{https://huggingface.co/datasets/Brianferrell787/financial-news-multisource}} \cite{Ferrell2025FinancialNews}. The official CLEF Task 3 evaluation set is drawn from the CLEF Task 3 Hugging Face\footnote{\href{https://huggingface.co/datasets/TheFinAI/CLEF_Task3_Trading}{https://huggingface.co/datasets/TheFinAI/CLEF\_Task3\_Trading}} dataset \cite{FinMMEvalTask3Overview2026}, which provides daily closing prices and pre-linked news for the testing phase. Volume data taken from Yahoo Finance is then injected into this data split to compute volume-based features. 

To isolate asset-specific information, news articles are filtered using regular expressions. We retain articles that contain at least two mentions of the target asset's keywords (\emph{Tesla}/\emph{TSLA} or \emph{Bitcoin}/\emph{BTC}), or at least one mention within the first 100 words (the lede). Following standard journalistic structure, an early mention likely indicates the asset is the primary subject of the article. This constraint is imposed to prevent the sentiment classifier from scoring noisy texts where the target asset is not a major detail. 

The text is then normalized only for the purposes of filtering and removing duplicates by converting to lowercase, removing punctuation, and stripping excess whitespace. We remove duplicates from the dataset by computing an MD5 hash of this normalized text; when duplicates are found, priority is given to records containing explicit publisher metadata. To preserve the semantic nuance, financial metrics, and tone required for accurate LLM inference, we retain the original unnormalized text for the sentiment classification step. Finally, we structure this input text by prepending available publisher information (e.g., \texttt{[Publisher: Name] Article Text}) to provide structured context for the sentiment model.



\paragraph{Temporal Alignment and Aggregation.}
All article timestamps are in UTC. To prevent look-ahead bias on TSLA, articles published after 16:00 UTC (i.e., after the U.S. equity market close) are shifted and linked to the following trading day. For BTC, which trades 24/7, we instead anchor each trading day to its UTC calendar date and link every article timestamped on day $t$ to day $t$'s closing price. Because trading decisions for day $t{+}1$ are conditioned strictly on information available at or before the close of day $t$, no look-ahead is introduced in either setting.

Following the article-level sentiment scoring (detailed in Section \ref{se:sentiment}), the data is aggregated on a daily basis. For each trading day, we calculate the mean sentiment score across all relevant articles and record the final closing price and volume. Lastly, standard percentage returns and log returns are computed to serve as inputs for the DRL environment.

\paragraph{Data Splits.}
We partition the augmented dataset chronologically into three distinct splits: training (prior to January 1, 2023), validation (January 1, 2023 to August 1, 2024), and testing (the official CLEF Task 3 evaluation period). This allocation yields 324 validation days for TSLA and 457 for BTC. To eliminate cold-start artifacts from our rolling-window technical indicators (e.g., EMA$_{50}$), the first 60 days of the training set are used only for indicator initialization and subsequently discarded. Similarly, to ensure the test set receives fully warmed-up features on its very first day, we prepend the final 60 days of the validation set to the test data prior to computing the indicators, truncating these warmup rows before evaluation.

\subsubsection{Market Regime Analysis}
\begin{figure}[h]
    \centering
    \caption{Asset prices (in log scale) across time and data splits.}
    \begin{subfigure}{\linewidth}
        \centering
        \caption{TSLA prices (in log scale) across time and data splits.}
        \includegraphics[width=\linewidth]{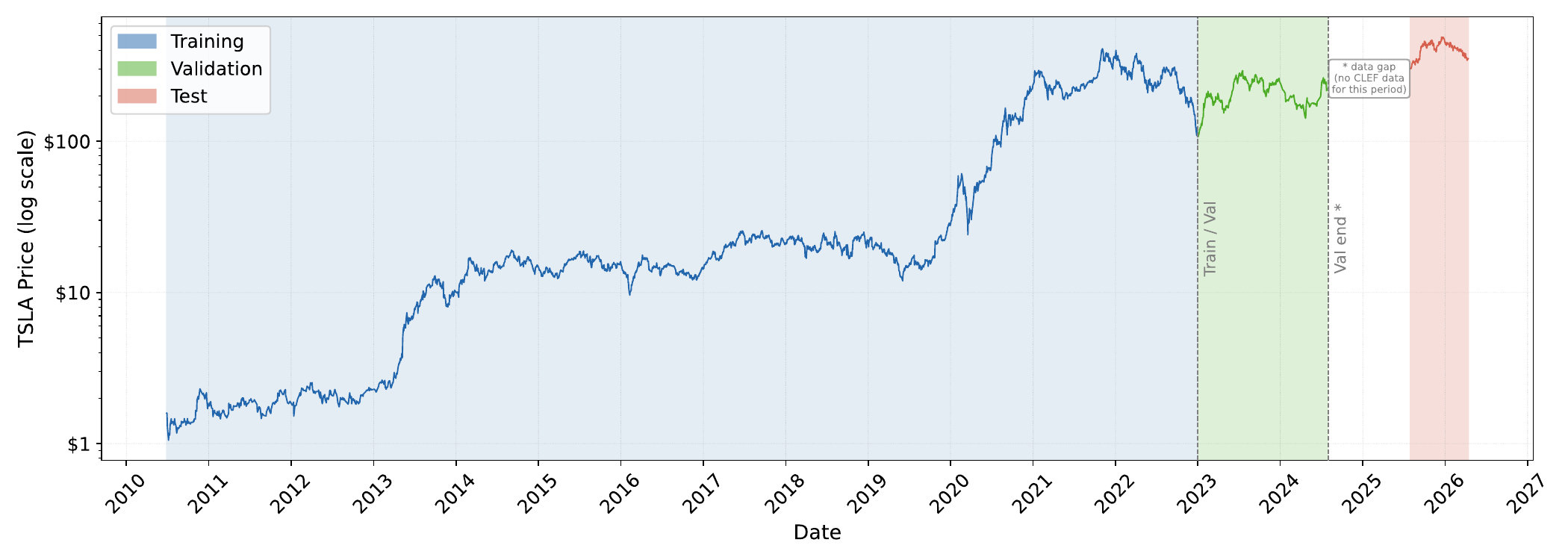}
        \label{fig:tsla_prices}
    \end{subfigure}
    \vspace{-2em}
    \begin{subfigure}{\linewidth}
        \centering
        \caption{BTC prices (in log scale) across time and data splits.}
        \includegraphics[width=\linewidth]{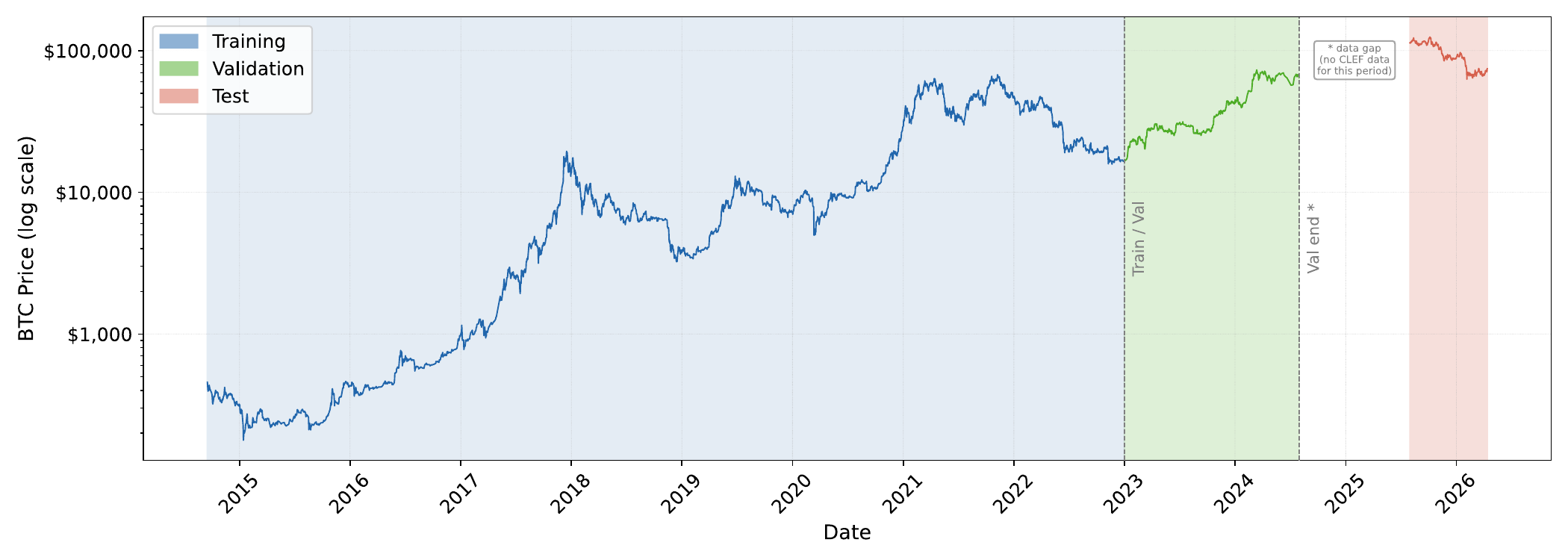}
        \label{fig:btc_prices}
    \end{subfigure}
    \label{fig:asset_prices}
\end{figure}

The three temporal splits differ substantially in their market dynamics, which is central to interpreting the generalisation results in Section \ref{se:results}. Figures \ref{fig:tsla_prices} and \ref{fig:btc_prices} show the prices (in log scale) over time for TSLA and BTC respectively. Identified are also the training, validation and test splits. Table \ref{tab:regimes} summarises statistics for each split and asset.

\begin{table}[h]
\caption{Market statistics across data splits. Ann.\ Ret., Ann.\ Vol., and SR are annualised; Cumul.\ Ret.\ is the buy-and-hold return over the full split. The Sharpe ratio (SR) uses a zero risk-free rate. Article counts for the test split are provided live by the competition.}
\label{tab:regimes}
\centering
\footnotesize
\begin{tabular}{l|l|l|r|r|r|r|r|r}
    \toprule
    \toprule
    \textbf{Asset} & \textbf{Split} & \textbf{Period} & \textbf{Days} & \textbf{Articles} & \textbf{Cumul.\ Ret.\ (\%)} & \textbf{Ann.\ Ret.\ (\%)} & \textbf{Ann.\ Vol.\ (\%)} & \textbf{SR} \\
    \midrule
    \midrule
    \multirow{3}{*}{TSLA}
        & Training & 2010--2022 & 2835 & 57{,}741 & 7{,}634.2 & 47.2 & 60.3 & 0.94 \\
        & Validation & 2023--2024 & 324 & 7{,}089 &   105.9 & 75.4 & 60.2 & 1.24 \\
        & Test$\dagger$ & 2025-- & 256 & ---   &    16.5 & 16.2 & 34.9 & 0.61 \\
    \midrule
    \multirow{3}{*}{BTC}
        & Training & 2014--2022 & 2664 & 29{,}307 & 3{,}518.3 & 40.4 & 65.0 & 0.85 \\
        & Validation & 2023--2024 & 457 &  2{,}861 &   288.7 & 111.4 & 43.3 & 1.95 \\
        & Test$\dagger$ & 2025-- & 256 & ---  &  $-$34.3 & $-$33.8 & 39.1 & $-$0.87 \\
    \bottomrule
    \multicolumn{8}{l}{$\dagger$ Measured to April 12th 2026; test set is continuously updated.}
\end{tabular}
\end{table}

As shown in Figure \ref{fig:asset_prices}, the \textit{training period} encompasses multiple complete market cycles for both assets. TSLA's history includes the rapid growth phase of 2020--2021 (share price rising from \$30 to a peak of \$407), followed by a severe correction throughout 2022 ($-65\%$). BTC experienced the 2017 speculative bubble and subsequent crash throughout 2018 ($-83\%$), the COVID-19 drawdown in March 2020, the 2021 all-time high near \$69,000, and the 2022 bear market ($-64\%$ year-over-year). The resulting high annualised volatility (60--65\%) (see Table~\ref{tab:regimes}) and long-time horizons expose agents to a diverse set of market conditions, providing broad coverage of both trending and mean-reverting regimes.

The \textit{validation period} (2023--2024) coincides with a sustained bull market recovery for both assets. BTC rebounded from its 2022 lows and reached new all-time highs, producing a cumulative return of 288.7\% over the period. TSLA also recovered strongly, yielding 105.9\%. The uniformly positive and high-momentum nature of the validation environment means that model selection based on validation Sharpe ratio may favour policies that have learned to hold long positions persistently, potentially at the expense of robustness to downturns. 

The \textit{test period} (from August 2025) presents a divergent regime for the two assets. TSLA continued its general upward trend at markedly reduced volatility (34.9\% annualised versus 60.2\% in validation), yielding a $16.5\%$ cumulative return (annualised SR of $0.61$), well below the validation SR of $1.24$. BTC, by contrast, entered a bear phase with a $-34.3\%$ cumulative return (annualized SR of $-0.87$), representing a sharp distributional shift from the $288.7\%$ bull market validation environment (SR of $1.95$). This regime divergence is the primary driver of the generalisation gap discussed in Section \ref{se:results}, particularly for BTC, where policies selected for strong bull market performance were exposed to sustained downward price pressure not well represented in the validation period.

\subsection{Feature Representation}
\label{se:features}

Table \ref{tab:features} lists the 15-dimensional state vector presented to each RL agent on each trading day. All technical indicators are normalized to be approximately zero-centered. We describe the rationale for each feature group below. 

\begin{table}[h]
    \centering
    \small
    \caption{State feature vector (14 market dimensions + previous position = 15 total inputs).}
    \label{tab:features}
    \begin{tabular}{c|c|l|l|l|l}
        \toprule
        \toprule
        & \textbf{\#} & \textbf{Category} & \textbf{Feature} & \textbf{Formula} & \textbf{Range}\\
        \midrule
        \midrule
        & 1 & Price & LogReturn & $\log(c_t/c_{t-1})$ & $\mathds{R}$\\
        \midrule
        & 2 & Agent State & Position & $p_{t-1}$ & $\{-1, 0, 1\}$\\
        \midrule
        & 3 & \multirow{6}{*}{Calendar} & DoW\_sin & $\sin(2\pi\cdot\text{DoW}/7)$ & $[-1,1]$\\
        & 4 & & DoW\_cos & $\cos(2\pi\cdot\text{DoW}/7)$ & $[-1,1]$\\
        & 5 & & DoM\_sin & $\sin(2\pi\cdot\text{DoM}/31)$ & $[-1,1]$\\
        & 6 & & DoM\_cos & $\cos(2\pi\cdot\text{DoM}/31)$ & $[-1,1]$\\
        & 7 & & MoY\_sin & $\sin(2\pi\cdot\text{MoY}/12)$ & $[-1,1]$\\
        & 8 & & MoY\_cos & $\cos(2\pi\cdot\text{MoY}/12)$ & $[-1,1]$\\
        \midrule
        \multirow{6}{*}{\rotatebox{90}{\scriptsize Technical Indicators}}
          & 9  & \multirow{2}{*}{Trend}    & EMA$_{10}$ & $\text{EMA}_{10}/c_t-1$ & $\mathds{R}$\\
          & 10 &                            & EMA$_{50}$ & $\text{EMA}_{50}/c_t-1$ & $\mathds{R}$\\
        \cline{2-6}
          & 11 & \multirow{2}{*}{Momentum} & RSI$_{14}$ & $\text{RSI}_{14}/50-1$ & $\mathds{R}$\\
          & 12 &                            & MACD & $(\text{EMA}_{12} - \text{EMA}_{26} - \text{signal})/c_t$ & $\mathds{R}$\\
        \cline{2-6}
          & 13 & Volatility & BollingerB & $(c_t-\text{SMA}_{20})/(2\sigma_{20})$ & $\mathds{R}$\\
        \cline{2-6}
          & 14 & Volume & LogVolumeChange & $\log(V_t/V_{t-1})$ & $\mathds{R}$\\
        \midrule
        & 15 & Sentiment & Sentiment & $p_+ - p_-$ (LLaMA 3.2 1B) & $[-1,1]$\\
        \bottomrule
    \end{tabular}
\end{table}

\paragraph{Price (LogReturn).}
The daily log return, computed as $\log(c_t/c_{t-1})$, provides a stationary, scale-invariant measure of the asset's immediate price movement over the last 24 hours. In quantitative finance, log returns are preferred over simple percentage returns because they are time-additive and symmetric, meaning positive and negative movements of the same magnitude cancel each other out mathematically. This feature anchors the agent to the most recent, immediate price action before the slower moving averages and momentum oscillators have time to react.

\paragraph{Agent State (Position).}
The agent's position from the previous time step, $p_{t-1} \in \{-1, 0, 1\}$, represents its internal state. Providing the agent with its own prior action is critical in environments with transaction fees. Without knowing its current portfolio allocation, the agent cannot evaluate the penalty of executing a new trade versus holding an existing position. Including this feature enables the DRL policy to learn "inertia", avoiding excessive trading churn and only triggering a state change when the expected market return exceeds the $0.2\%$ transaction cost barrier.

\paragraph{Calendar Features (Cyclical Encodings).}
The six sinusoidal encodings in Table~\ref{tab:features} represent day-of-week (DoW), day-of-month (DoM), and month-of-year (MoY) as sine/cosine pairs, providing the agent with a continuous, cycle-aware representation of calendar time. Financial markets exhibit well-documented intra-week and intra-month seasonality: the day-of-week effect \cite{French1980Weekend} and turn-of-the-month patterns in equity returns are two examples. Encoding calendar variables as sine/cosine pairs rather than as raw integers preserves the cyclic continuity and avoids imposing an ordinal ranking on categorical time features.  

\subsubsection{Technical Indicators}

\paragraph{Trend Indicators (EMA$\mathbf{_{10}}$, EMA$\mathbf{_{50}}$).}
Exponential Moving Averages (EMAs) assign exponentially decaying weights to past prices, making them more responsive to recent changes than simple moving averages \cite{Murphy1999Technical}. EMA$_{10}$ captures short-term momentum over approximately two trading weeks, while EMA$_{50}$ reflects the medium-term trend over roughly 2.5 months. Expressing each EMA relative to the current close ($\text{EMA}/c_t-1$) produces a scale-invariant feature that is directly comparable across assets with vastly different price levels (e.g. BTC at \$60,000 versus TSLA at \$300). The combination of a fast and a slow EMA also encodes an implicit cross-signal: when EMA$_{10} >$ EMA$_{50}$, price is in a short-term uptrend relative to the medium-term baseline, a classical bullish signal \cite{Brock1992SimpleTechnical}. 

\paragraph{Momentum Indicators (RSI$\mathbf{_{14}}$, MACD).} 
The Relative Strength Index (RSI$_{14}$) \cite{Wilder1978NewTechnical} is a bounded oscillator computed as the ratio of average gains to average losses over a 14-day lookback. We re-centre it from $[0,100]$ to $[-1,1]$ via $\text{RSI}_{14}/50-1$, so that the neutral level is zero. Values above $0.4$ (RSI$_{14}>70$) signal overbought conditions associated with potential reversals, while values below $-0.4$ (RSI$_{14}<30$) signal oversold conditions, providing a mean-reversion counterpoint to the trend-following EMA features. 

Moving Average Convergence Divergence (MACD) \cite{Appel2005Technical} measures the histogram between the MACD line (EMA$_{12}-$EMA$_{26}$) and its 9-day exponential signal line, normalised by the closing price to remove scale dependence. The MACD histogram captures trend acceleration and deceleration, providing a leading signal for potential reversals before they are evident in price alone. Normalisation by $c_t$ ensures the feature remains stationary across different price regimes. 

\paragraph{Volatility Indicator (BollingerB).}
BollingerB \cite{Bollinger2001Bollinger} positions the current price within a band defined as two standard deviations above and below a 20-day simple moving average (SMA$_{20}$):
\begin{equation*}
    \text{BollingerB} = \frac{c_t-\text{SMA}_{20}}{2\sigma_{20}}.
\end{equation*}
Unlike RSI, whose thresholds are fixed regardless of volatility regimes, BollingerB adapts its thresholds dynamically to the prevailing volatility regime, making it particularly informative for assets like BTC that alternate between low-volatility consolidation periods and high-volatility breakouts. Values near $\pm 1$ indicate price at the band extremes and are commonly associated with reversal or continuation signals depending on the volume context.

\paragraph{Volume Indicator (LogVolChange).}
The log ratio of consecutive daily volumes $\log(V_t/V_{t-1})$ captures day-over-day shifts in market participation. Trading volume is widely recognized as a confirmation signal: price movements accompanied by abnormally high volume are considered more reliable indicators of trend continuation than those occurring on thin volume \cite{Lo2000Foundations}. Using a log ratio rather than raw volume renders the feature invariant to the absolute scale of trading activity, which differs by orders of magnitude between TSLA (equity market) and BTC (24/7 cryptocurrency market).

\subsection{Sentiment Analysis}
\label{se:sentiment}
As shown in Figure \ref{fig:system}, we score daily news sentiment using the LLaMA 3.2 1B Instruct model \cite{dubey2024llama3} in a zero-shot setting. We selected this model because modern instruction-tuned LLMs demonstrate strong zero-shot reasoning capabilities that often match or exceed older domain-specific models (e.g., FinBERT), while its 1B parameter footprint ensures computational efficiency for daily inference. To capture the core market signal and avoid diluting the sentiment with irrelevant body content, each article is truncated to its first 300 words, in order to focus on the headline and the lede. The truncated text is then wrapped in a structured conversational prompt:

\begin{center}
\begin{tcolorbox}[colback=gray!5, colframe=black!40, boxrule=0.5pt, arc=2pt, left=5pt, right=5pt, top=5pt, bottom=5pt, width=0.95\linewidth]
\textbf{[System]} \textit{You are a financial sentiment classifier for \{asset\}. Respond with exactly one word: positive, negative, or neutral.}\\[1ex]
\textbf{[User]} \textit{\{article text\}}
\end{tcolorbox}
\end{center}

Rather than relying on standard autoregressive text generation, which can be
computationally expensive and subject to formatting hallucinations, we employ
a log-probability extraction approach. We perform a single forward pass through
the model and extract the logits at the final token position corresponding to
the model's output. To avoid brittleness to a single tokenization, for each
class label we enumerate the BPE token-ID variants with and without a leading
space, and with lower- and title-case (e.g., \texttt{positive},
\texttt{\textvisiblespace positive}, \texttt{Positive},
\texttt{\textvisiblespace Positive}). A softmax is then applied strictly over
this union of token IDs, and the resulting probabilities are summed per class
to yield $P(\text{positive})$, $P(\text{negative})$, and $P(\text{neutral})$.

The final sentiment score for a given article is calculated as the difference between the aggregated probabilities of the positive and negative classes:
\begin{equation}
    Sentiment = P(\text{positive}) - P(\text{negative})
\end{equation}

Although the neutral class probability $P(\text{neutral})$ does not appear explicitly in this equation, it plays an important structural role. Because the values are derived from a softmax distribution where $P(\text{positive}) + P(\text{negative}) + P(\text{neutral}) = 1$, a high probability assigned to the neutral class restricts the available probability mass for the polar classes. Therefore, a strong neutral prediction acts as an automatic confidence weight, dampening the final sentiment score and driving it toward zero. This formulation yields a bounded continuous score in the range $[-1, 1]$.

\subsection{Alpha Reward}
An important design choice is the \emph{alpha reward}: instead of rewarding absolute portfolio log-return, we reward excess return over the market:
\begin{equation}
    r_t = \log\left(\frac{V_t}{V_{t-1}}\right) - \log\left(\frac{c_t}{c_{t-1}}\right).
\end{equation}
Under a passive buy-and-hold strategy with no costs, every step yields $r_t=0$. The agent is therefore incentivised to outperform the market rather than merely profit from rising prices, directly aligning the training objective with the evaluation criterion. Model selection on the validation set uses the checkpoint with the highest Sharpe ratio computed on raw portfolio values, matching the task's evaluation metric.

\paragraph{Theoretical Properties of the Alpha Reward.}
Beyond its interpretation as market-relative performance, the alpha reward has three properties worth noting. Let $m_t=\log(c_t/c_{t-1})$ denote the market log-return, so that $r_t=\log(V_t/V_{t-1})-m_t$.

\emph{1. Cumulative alpha identity.} Over an undiscounted episode of length $T$, the per-step rewards telescope to:
\begin{equation}
\sum_{t=1}^{T} r_t \;=\; \log\!\left(\frac{V_T}{V_0}\right) - \log\!\left(\frac{c_T}{c_0}\right) \;=\; \log(\alpha_T),
\end{equation}

where $\alpha_T=(V_T/V_0)/(c_T/c_0)$ is the terminal outperformance ratio over buy-and-hold. The return-to-go $G_t$ is therefore the log-alpha of the strategy, directly aligning the training signal with the CLEF Task~3 evaluation criterion.

\emph{2. Equivalent optima to the raw reward.} Under the price-taker assumption (see Section~\ref{se:background}), the market return sequence $\{m_t\}$ is independent of the policy $\pi_\theta$. For any discount factor $\gamma\in[0,1]$, the alpha-reward objective therefore differs from the raw log-return objective only by a policy-independent constant $\mathds{E}[\sum_t \gamma^{t-1} m_t]$, and the two share the same optimal policies. This is a weaker condition than the potential-based shaping of~\cite{Ng1999PolicyInvariance}, but suffices here because $m_t$ depends only on the state transition.

\emph{3. Variance reduction.} Although the optima coincide, the learning dynamics do not. For a long position with no transaction cost, $\log(V_t/V_{t-1}) = m_t$, so the raw reward has variance $\mathrm{Var}(m_t)$ whereas the alpha reward has variance zero; the agent receives a clean signal about its deviation from buy-and-hold rather than a noisy signal dominated by market moves. More generally, subtracting $m_t$ acts as a control variate on the reward, reducing gradient-estimator variance with no bias cost. On assets with high market volatility, TSLA and BTC both exceed $60\%$ annualised training volatility (see Table~\ref{tab:regimes}), this variance reduction is substantial.

\subsection{DRL Algorithms}
\label{se:DRL}

We compare four DRL algorithms that span the main families of policy- and value-based approaches. 

\paragraph{Policy Gradient (PG).}
\label{se:PG}

PG \cite{Williams1992REINFORCE} estimates the policy gradient as $\nabla_\theta J = \mathds{E}[\nabla_\theta\log\pi_\theta(a|s)\cdot G]$ using Monte-Carlo returns $G$ normalized within each episode for variance reduction. An entropy bonus $\beta H(\pi)$ discourages premature collapse to a deterministic policy. We parameterized $\pi_\theta$ as an LSTM network to capture temporal dependencies across the trading sequence. 

\paragraph{Proximal Policy Optimization (PPO).}
\label{se:PPO}

PPO \citep{Schulman2017PPO} improves stability over vanilla PG by clipping the probability ratio:
\begin{equation}
    r_t(\theta) = \frac{\pi_\theta(a|s)}{\pi_{\theta_\text{old}}(a|s)}    
\end{equation}
within $[1-\varepsilon, 1+\varepsilon]$. We use an LSTM actor-critic network with Generalized Advantage Estimation (GAE-$\lambda$) and truncated back-propagation through time (TBPTT) to manage gradient flow over long sequences.

\paragraph{Deep Q-Learning (DQL).}
\label{se:DQL}

DQL \cite{MnihDQN} learns an action-value function $Q(s,a;\theta)$ via temporal-difference learning with experience replay and a target network. We adopt the double DQL \cite{vanHasselt2015DoubleDQL} update to reduce overestimation bias. The network is a feed-forward architecture (FFNN) since the replay buffer breaks temporal ordering, making LSTM less appropriate.

\paragraph{Deep Deterministic Policy Gradient (DDPG).}
\label{se:DDPG}

DDPG \cite{Lillicrap2016continuous} extends DQL to continuous action spaces with a deterministic actor $\mu_\theta(s)$ and a critic $Q_\phi(s,a)$. We discretize the continuous output to $\{-1,0,1\}$ via thresholds: $a<-0.33 \Rightarrow$ \textsc{short}, $a > 0.33 \Rightarrow$ \textsc{long}, else \textsc{flat}. The threshold $1/3$ partitions the actor's output range $[-1,1]$ into three equal-width intervals, giving each discrete action equal prior volume and avoiding an implicit bias toward any of the three positions. Exploration is driven by decaying Ornstein-Uhlenbeck noise. Like DQL, the actor and critic are FFNNs due to the use of a replay buffer.

\section{Experimental Setting}
\label{se:experimental}

\subsection{Hyperparameter Optimisation}

We use Ray Tune to run random search over 180 configurations per algorithm-asset pair (8 combinations) on a 192-core CPU cluster (Trillium, Digital Research Alliance of Canada), running all 180 trials in parallel. The search spaces are summarized in Table \ref{tab:hyperparameter}. All trials train for up to 2000 episodes with validation every 50 episodes; a trial terminates early if the validation SR does not improve for 10 consecutive evaluations. The configuration with the highest validation performance is selected.

\begin{table}[h]
    \centering
    \small
    \caption{Hyperparameter search space.}
    \label{tab:hyperparameter}
    \begin{tabular}{l|l}
        \toprule
        \toprule
        \textbf{Parameter} & \textbf{Values}\\
        \midrule
        \midrule
        hidden\_dim & $\{64,128,256\}$\\
        num\_layers & $\{1,2,3\}$\\
        $\gamma$ & $\{0.95,0.99\}$\\
        lr & $[10^{-4},10^{-2}]$\\
        entropy coeff (PG/PPO) & $[10^{-3}, 5\cdot10^{-2}]$\\
        clip $\varepsilon$ (PPO) & $\{0.1,0.2,0.3\}$\\
        $\tau$ (DDPG) & $\{0.001,0.005,0.01\}$\\
        \bottomrule
    \end{tabular}
\end{table}

\subsection{Training Details}

We implement a custom trading environment modeled after the OpenAI Gym interface. Each training episode starts at a randomly sampled day (with at least 100 days remaining) forcing the agent to encounter diverse market conditions rather than memorizing a fixed sequence. During evaluation, the episode always begins from the first day of the period.

Hyperparameter optimization uses a fixed train/validation split: agents are trained on data prior to January 1st 2023 and evaluated on data between January 2023 and August 2024. Once the best configuration is identified, each model retrains from scratch on the same training split (pre-2023) with early stopping based on the validation SR (patience 10). The best validation SR checkpoint from this final run is saved and used for all reported test results. We apply L2 regularization, with weight decay $10^{-4}$, to all Adam optimizers and gradient clipping norm $1.0$. 

Transaction costs of $0.2\%$ are applied during training only, causing the policy to internalize an inertia incentive, while evaluation is cost-free to exactly match the CLEF Task~3 submission protocol, which itself does not deduct execution costs. We acknowledge that this difference in transaction costs application reflects the task's evaluation convention and that realistic deployment would require re-introducing costs at inference; we view this as a limitation of external validity rather than of the method. 

The alpha reward and random episode start are enabled during training only. Validation and test evaluation always start from the first day of the respective period. Additionally, model selection uses the Sharpe ratio computed on the raw portfolio value series (matching how test SR is defined). Our test backtest mirrors the CLEF live-submission protocol (daily actions, cost-free, identical metric definitions); the only procedural difference is that we evaluate on a fixed snapshot of the test window rather than the continuously-updated leaderboard window.

\section{Results}
\label{se:results}

Table \ref{tab:combined_results} reports the final performance of each trained agent on the CLEF Task 3 test set. The TSLA buy-and-hold baseline for the test period is $16.45\%$ (SR $0.61$, MDD $29.93$); the BTC baseline is $-34.27\%$ (SR $-0.87$, MDD $49.72$). 

\begin{table}[h]
    \caption{Test set return and risk metrics on CLEF Task 3. Test CR = cumulative return (\%), Test SR = Sharpe ratio, Val CR and Val SR are measured on the validation set, MDD = maximum drawdown (\%), DV = daily volatility (\%), AV = annualised volatility (\%). Model selection uses the highest Val SR.}
    \label{tab:combined_results}
    \centering
    \small
    \begin{tabular}{l|l|r|r|r|r|r|r|r}
        \toprule
        \toprule
        \textbf{Asset} & \textbf{Model} & \textbf{Test CR (\%)} & \textbf{Test SR} & \textbf{Val CR (\%)} & \textbf{Val SR} & \textbf{MDD (\%)} & \textbf{DV (\%)} & \textbf{AV (\%)}\\
        \midrule
        \midrule
        \multirow{5}{*}{TSLA}
            & B\&H          & $16.45$          & $0.61$          & $105.94$          & $1.24$          & $29.93$          & $2.19$          & $34.80$          \\
            \cdashline{2-9}
            & PG            & $13.96$          & $0.55$          & $101.76$          & $1.21$          & $29.93$          & $2.19$          & $34.74$          \\
            & PPO           & $2.54$           & $0.24$          & $169.50$          & $1.75$          & $28.38$          & $\mathbf{2.01}$ & $\mathbf{31.93}$ \\
            & DQL           & $52.62$          & $1.38$          & $\mathbf{830.30}$ & $\mathbf{3.29}$ & $35.76$          & $2.19$          & $34.69$          \\
            & \textbf{DDPG} & $\mathbf{54.96}$ & $\mathbf{1.44}$ & $262.10$          & $2.03$          & $\mathbf{19.20}$ & $2.15$          & $34.09$          \\
        \midrule
        \multirow{5}{*}{BTC}
            & B\&H          & $-34.27$         & $-0.87$         & $\mathbf{288.69}$ & $1.95$          & $49.72$          & $2.46$          & $39.00$          \\
            \cdashline{2-9}
            & PG            & $-14.53$         & $-0.33$         & $109.33$          & $1.43$          & $\mathbf{29.98}$ & $\mathbf{1.98}$ & $\mathbf{31.51}$ \\
            & PPO           & $-33.80$         & $-0.85$         & $190.07$          & $1.58$          & $59.18$          & $2.46$          & $39.00$          \\
            & DQL           & $-23.22$         & $-0.55$         & $276.71$          & $\mathbf{2.11}$ & $37.41$          & $2.25$          & $35.72$          \\
            & \textbf{DDPG} & $\mathbf{1.58}$  & $\mathbf{0.23}$ & $259.53$          & $1.87$          & $35.30$          & $2.40$          & $38.12$          \\
        \bottomrule
    \end{tabular}
\end{table}

For TSLA, as detailed in Table \ref{tab:combined_results}, both DDPG (CR $54.96\%$, SR $1.44$) and DQL (CR $52.62\%$, SR $1.38$) outperform the buy-and-hold (B\&H) baseline (CR $16.45\%$, SR $0.61$), demonstrating that RL agents can learn to time the TSLA market effectively. PG (CR $13.96\%$, SR $0.55$) falls just below the baseline, and PPO (CR $2.54\%$, SR $0.24$) trails behind significantly despite a large validation return (Val CR $169.50\%$). Examining the risk metrics, DDPG again stands out with the lowest maximum drawdown (MDD $19.20\%$), well below the buy-and-hold drawdown (MDD $29.93\%$), while DQL incurs a higher drawdown (MDD $35.76\%$) despite its competitive return. PG precisely matches the buy-and-hold drawdown (MDD $29.93\%$), reflecting a policy that largely tracks the market.

BTC presents a substantially harder generalisation challenge. The B\&H baseline itself has a Sharpe ratio of $-0.87$ and a maximum drawdown of $49.72\%$, reflecting a severe bear market in the test period. All four models outperform buy-and-hold on cumulative return and Sharpe ratio. DDPG is the only model to achieve a positive return (CR $1.58\%$, SR $0.23$, MDD $35.30\%$), significantly outperforming buy-and-hold (CR $-34.27\%$, SR $-0.87$, MDD $49.72\%$). PG (CR $-14.53\%$, SR $-0.33$, MDD $29.98\%$) is the second-best, markedly reducing both losses and drawdown relative to the baseline. DQL ranks third (CR $-23.22\%$, SR $-0.55$, MDD $37.41\%$), while PPO (CR $-33.80\%$, SR $-0.85$, MDD $59.18\%$) ranks last, still outperforming buy-and-hold in terms of returns but incurring a substantially larger maximum drawdown, effectively replicating the market trajectory without the capital preservation benefit seen in the other models. DQL records the highest validation SR on BTC ($2.11$) yet finishes with a test SR of $-0.55$, illustrating how strong bull market validation performance can fail to generalise to a bear-market test regime.

\subsection{Algorithm Analysis}

DDPG is the standout algorithm, ranking first on both assets in return and Sharpe ratio. Its continuous-action actor-critic architecture, combined with a replay buffer and soft target updates appears to provide a more stable optimisation landscape under the alpha reward signal than recurrent on-policy methods (PG and PPO) or the purely off-policy DQL. On TSLA, DQL matches DDPG closely in return ($52.62\%$ vs $54.96\%$) but with a substantially larger drawdown ($35.76\%$ vs. $19.20\%$), suggesting riskier positioning from DQL. On BTC, no recurrent on-policy method (PG, PPO) achieves a positive return, and PPO nearly replicates the buy-and-hold trajectory with a higher drawdown. 

\begin{figure}[h]
    \centering
    \caption{DDPG vs B\&H portfolio values on the TSLA and BTC test sets.}
    \begin{subfigure}{\linewidth}
        \centering
        \caption{DDPG vs B\&H portfolio values on the TSLA test set.}
        \includegraphics[width=\linewidth]{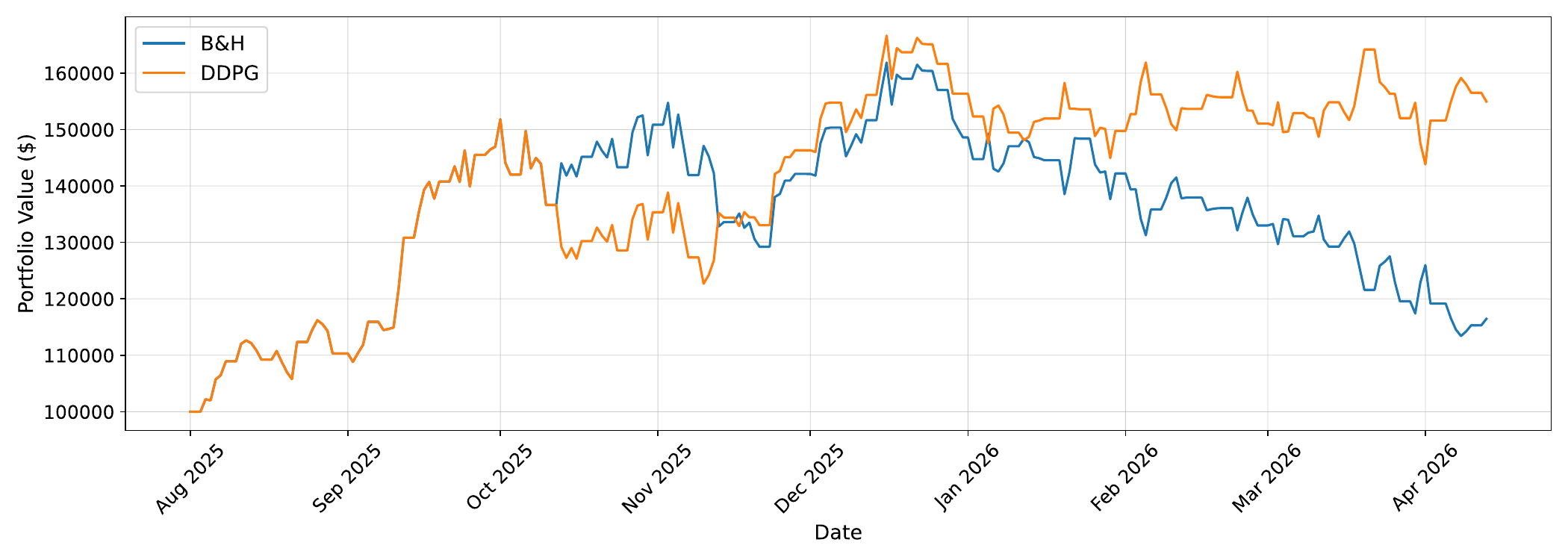}
        \label{fig:tsla_ddpg_portfolio}
    \end{subfigure}
    \vspace{-2em}
    \begin{subfigure}{\linewidth}
        \centering
        \caption{DDPG vs B\&H portfolio values on the BTC test set.}
        \includegraphics[width=\linewidth]{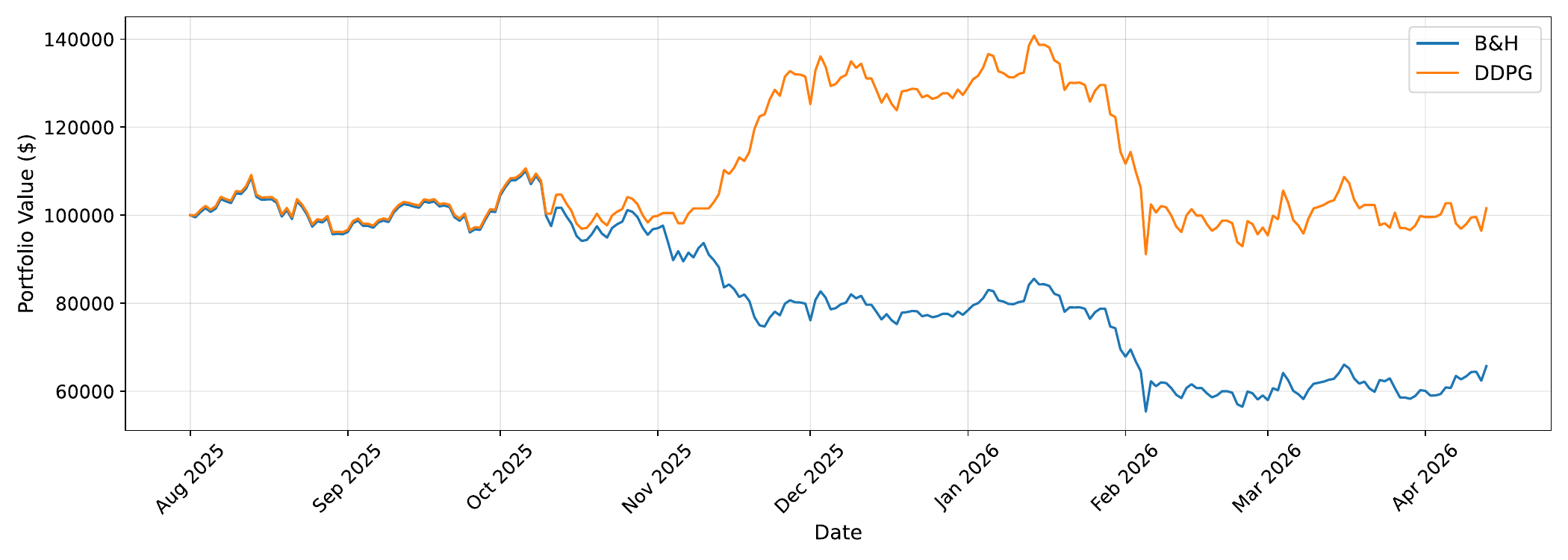}
        \label{fig:btc_ddpg_portfolio}
    \end{subfigure}
    \label{fig:ddpg_portfolio}
\end{figure}

Figures \ref{fig:tsla_ddpg_portfolio} and \ref{fig:btc_ddpg_portfolio} show the DDPG portfolio trajectory against the buy-and-hold on the test set for each asset.

On TSLA (Figure \ref{fig:tsla_ddpg_portfolio}), DDPG leads buy-and-hold throughout most of the test period. Both series rise together from August to December 2025, peaking over $\$160,000$ for buy-and-hold. After that peak, buy-and-hold retraces sharply as TSLA gives back gains, while DDPG preserves capital by reducing its long exposure, ending the period near $\$155,000$ versus $\$116,000$ for buy-and-hold. This sustained outperformance reflects DDPG's ability to time partial exits during the downswing, consistent with its best maximum drawdown of $19.20\%$.

On BTC (Figure \ref{fig:btc_ddpg_portfolio}), the two series track each other closely during the flat August--October 2025 phase, indicating that DDPG holds near-neutral positioning while the market drifts. From November 2025 onward, DDPG diverges sharply upward, reaching around $\$130,000$ by December, while buy-and-hold begins a sustained decline that eventually drops below $\$80,000$ by early 2026. The agent sustains a sharp drawdown in February 2026, pulling back to just over $\$90,000$ before stabilizing, but ultimately ends the period close to breakeven ($\$101,580$ vs $\$65,730$ for buy-and-hold). The contrast illustrates how DDPG's continuous actor effectively shifts from long to flat or short during sustained bearish periods, limiting participation in the worst of the drawdown. 

\begin{figure}[h]
    \centering
    \caption{DQL vs B\&H portfolio values on the TSLA and BTC test sets.}
    \begin{subfigure}{\linewidth}
        \centering
        \caption{DQL vs B\&H portfolio values on the TSLA test set.}
        \includegraphics[width=\linewidth]{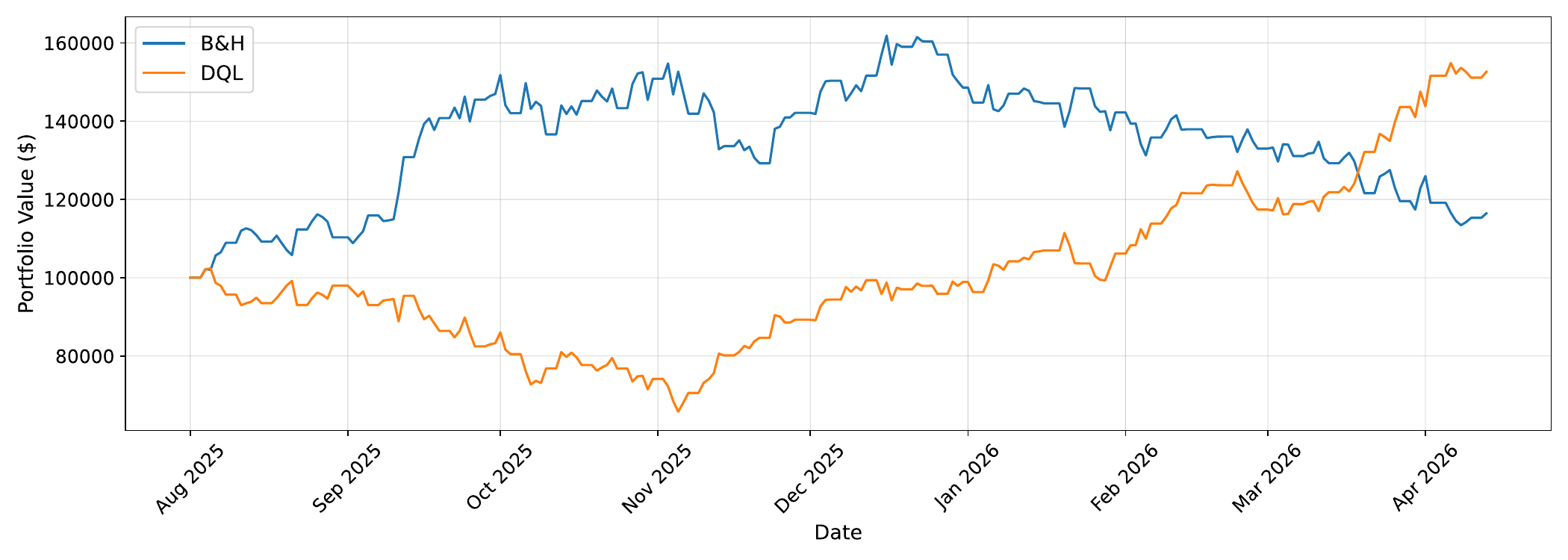}
        \label{fig:tsla_dql_portfolio}
    \end{subfigure}
    \vspace{-2em}
    \begin{subfigure}{\linewidth}
        \centering
        \caption{DQL vs B\&H portfolio values on the BTC test set.}
        \includegraphics[width=\linewidth]{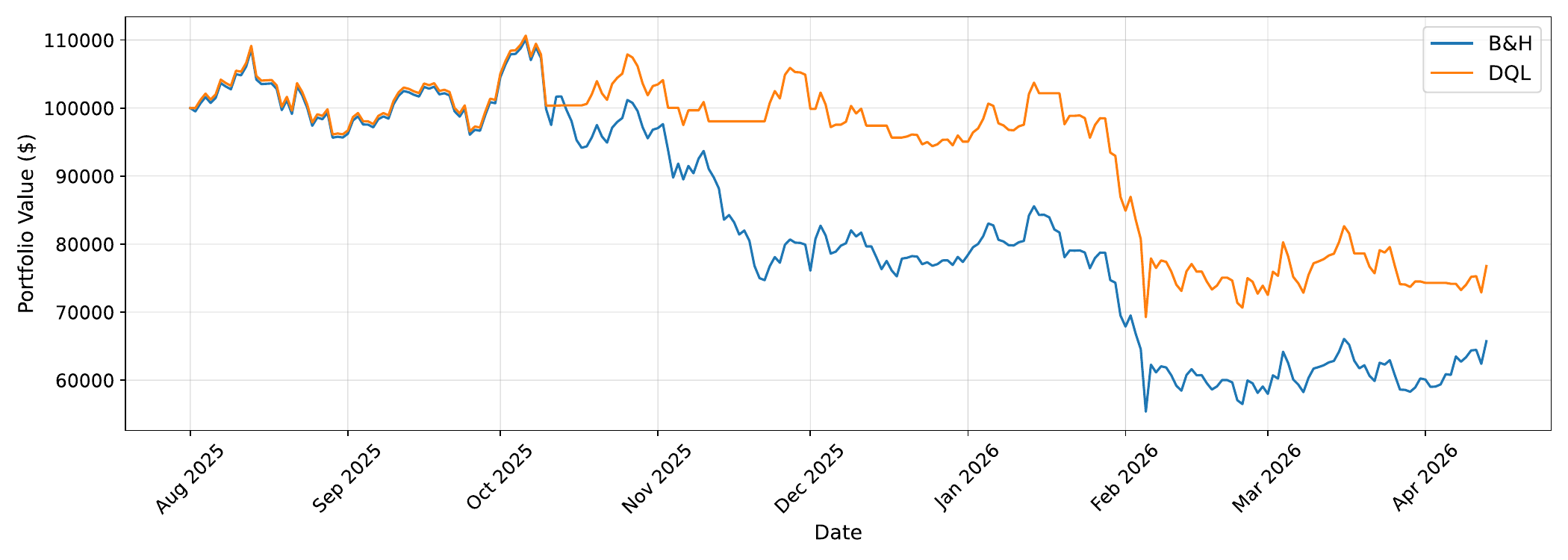}
        \label{fig:btc_dql_portfolio}
    \end{subfigure}
    \label{fig:dql_portfolio}
\end{figure}

DQL achieved the highest validation Sharpe ratio on both assets (TSLA: $3.29$, BTC: $2.11$) and was therefore selected as our submission model for the live CLEF competition rankings. We retain this choice despite DDPG's stronger offline backtest: our selection criterion was fixed a priori, and re-selecting the endpoint model based on a single realized test path would constitute selection bias. The gap between DQL's validation ranking and its backtest performance is instead reported as evidence of the regime-shift sensitivity analyzed in Section~\ref{se:validation_vs_test}. Figures \ref{fig:tsla_dql_portfolio} and \ref{fig:btc_dql_portfolio} show its test trajectories.

On TSLA (Figure \ref{fig:tsla_dql_portfolio}), DQL takes an initially incorrect short position while TSLA rises, falling to roughly $\$75,000$ by November 2025 and accounting for the agent's high maximum drawdown ($35.76\%$). It then reverses to a long position and steadily recovers, closing at around $\$152,000$, above buy and hold's $\$116,000$, for an overall gain of $52.62\%$. 

On BTC (Figure \ref{fig:btc_dql_portfolio}), DQL provides modest protection relative to buy-and-hold, staying closer to breakeven while B\&H steadily declines. Both suffer the February 2026 sell-off, after which DQL stabilizes around $\$75,000$ versus $\$65,000$ for buy-and-hold (CR $-23.22\%$ vs $-34.27\%$). The contrast between DQL's strong validation Sharpe ratio and its comparatively volatile test path highlights the regime-dependence of model selection discussed in Section~\ref{se:validation_vs_test}.

A persistent training artefact is the massive gap between training-episode returns and test returns: PPO reports training-period portfolio gains of $114{,}462\%$ on TSLA and $256{,}989\%$ on BTC. These inflated figures bear no direct relationship to the test performance and underscore the importance of Sharpe-based model selection over return-based selection. 

\subsection{Validation Sharpe Ratio vs. Test Performance}
\label{se:validation_vs_test}

The relationship between validation Sharpe ratio and test performance is asset-dependent. As shown in Table~\ref{tab:combined_results}, on TSLA, the two highest validation SR models (DQL: $3.29$, DDPG: $2.03$) are also the two best performers ($52.62\%$ and $54.96\%$ respectively), suggesting that on a moderately trending asset, validation Sharpe ratio provides a useful signal. On BTC, the pattern breaks: DQL leads on validation SR ($2.11$) but ranks third on the test set (CR $-23.22\%$, SR $-0.55$). DDPG, with a lower validation SR ($1.87$), is the only model to achieve a positive test return. This divergence arises because the BTC validation period (January 2023--August 2024) was a sustained bull market recovery, while the test period is a bear market; policies selected for strong upward momentum performance may not generalise to downturns. Overall, validation-based model selection eliminates the weakest checkpoints but remains subject to distributional shift between market regimes. 

A structural contributor to this gap is the use of a single temporal validation window that happened to coincide with a bull market for both assets. Walk-forward cross-validation, which would create multiple rolling train/validate splits that collectively span bull runs, bear markets, and sideways regimes, would yield a more regime-balanced model selection criterion. Implementing this technique would likely have reduced the observed validation-to-test gap. Ultimately, walk-forward cross-validation was not implemented due to time constraints as this method is very computationally demanding.

\section{Conclusion and Further Work}

We presented a DRL-based trading system for CLEF 2026 FinMMEval Task 3 that integrates LLM-derived news sentiment with technical indicators and calendar features in a custom trading environment. Our key contributions are the alpha reward formulation applied during training, paired with Sharpe-based model selection on the validation set, and random episode start, which together reduced overfitting to the training sequence and aligned the optimisation objective with the competitive criterion of beating buy-and-hold. DDPG was the most robust algorithm across both assets, achieving a $54.96\%$ cumulative return on TSLA (SR $1.44$, MDD $19.20\%$) and $1.58\%$ on BTC (SR $0.23$, MDD $35.30\%$). However, our live endpoint uses DQL, selected on validation Sharpe ratio without reference to the test-period data. On TSLA, both DDPG (CR $54.96\%$, SR $1.44$) and DQL (CR $52.62\%$, SR $1.38$) significantly outperformed the baseline, while PG and PPO fell short. On BTC, DDPG was the only model to achieve a positive return while PG substantially reduced losses relative to buy-and-hold. The relationship between validation SR and test performance is asset-dependent: on TSLA, the top two validation SR models were also the top two test performers; on BTC, regime shift from a bull validation period to a bear test period caused DQL (highest validation SR at $2.11$) to underperform DDPG (lower validation SR at $1.87$), exposing the limits of validation-based model selection under distributional shifts. 

Several directions remain for future work. First, replacing LLaMA 3.2 1B with the domain-specialised FinBERT \cite{Araci2019FinBERT} or a larger LLM may improve sentiment quality. Second, incorporating SEC 10-K/10-Q filings available in the task's daily context could provide longer-horizon signals. Third, walk-forward cross-validation over multiple regime-diverse validation windows would yield a more robust model selection criterion, reducing the bull market bias embedded in our single 2023--2024 validation split. Fourth, hierarchical or multi-timescale RL, learning both short-term momentum policies and longer-term regime-detection strategies, may help bridge the validation-to-test generalisation gap. Fifth, ensemble methods that combine the action outputs of multiple agents could smooth individual model noise and reduce the variance observed across algorithm-asset pairs. Finally, a systematic ablation study, selectively removing individual feature groups (technical indicators, sentiment scores, and calendar encodings) or replacing the LLM sentiment score with a simpler lexicon-based baseline, would quantify each component's marginal contribution to agent performance and clarify whether the observed gains are primarily driven by price-derived signals or by the news sentiment score.

\begin{acknowledgments}
This research was supported by funding from the Natural Sciences and Engineering Research Council of Canada (NSERC). We also gratefully acknowledge the Digital Research Alliance of Canada for providing the high-performance computing resources utilized throughout this project.
\end{acknowledgments}

\bibliography{references}

@book{Appel2005Technical,
    title     = {Technical Analysis: Power Tools for Active Investors},
    author    = {Appel, Gerald},
    publisher = {Financial Times Prentice Hall},
    year      = {2005}
}

@misc{Araci2019FinBERT,
      title={FinBERT: Financial Sentiment Analysis with Pre-trained Language Models}, 
      author={Dogu Araci},
      year={2019},
      eprint={1908.10063},
      archivePrefix={arXiv},
      primaryClass={cs.CL},
}

@book{Bollinger2001Bollinger,
    title     = {Bollinger on Bollinger Bands},
    author    = {Bollinger, John},
    publisher = {McGraw-Hill},
    year      = {2001}
}

@Article{Du2024DRL_CNN_Trading,
AUTHOR = {Du, Sha and Shen, Hailong},
TITLE = {A Stock Prediction Method Based on Deep Reinforcement Learning and Sentiment Analysis},
JOURNAL = {Applied Sciences},
VOLUME = {14},
YEAR = {2024},
NUMBER = {19},
ARTICLE-NUMBER = {8747},
URL = {https://www.mdpi.com/2076-3417/14/19/8747},
ISSN = {2076-3417},
DOI = {10.3390/app14198747}
}

@article{Brock1992SimpleTechnical,
    title   = {Simple Technical Trading Rules and the Stochastic Properties of
             Stock Returns},
    author  = {Brock, William and Lakonishok, Josef and LeBaron, Blake},
    journal = {Journal of Finance},
    volume  = {47},
    number  = {5},
    pages   = {1731--1764},
    year    = {1992}
}

@misc{Ferrell2025FinancialNews,
    author       = { Brian Ferrell },
    title        = { financial-news-multisource (Revision b509ef6) },
    year         = { 2025 },
    doi          = { 10.57967/hf/6432 },
    publisher    = { Hugging Face }
}

@inproceedings{FinMMEvalTask3Overview2026,
  title = {Overview of the {FinMMEval} 2026 Task 3: Financial Decision Making},
  author = {Zhuohan Xie and Lingfei Qian and Georgi Georgiev and Dimitar Dimitrov and Rania Elbadry and Fan Zhang and Xueqing Peng and Jimin Huang and Vanshikaa Jani and Yuyang Dai and Jiahui Geng and Yankai Chen and Ye Yuan and Haolun Wu and Yuxia Wang and Ivan Koychev and Veselin Stoyanov and Mingzi Song and Yu Chen and Steve Liu and Preslav Nakov},
  booktitle = {CLEF 2026 Working Notes},
  series = {CEUR Workshop Proceedings},
  year = {2026},
  month = {September 21--24},
  address = {Jena, Germany},
  publisher = {CEUR-WS.org},
}

@article{French1980Weekend,
  title   = {Stock Returns and the Weekend Effect},
  author  = {French, Kenneth R},
  journal = {Journal of Financial Economics},
  volume  = {8},
  number  = {1},
  pages   = {55--69},
  year    = {1980}
}

@misc{Jiang2017Portfolio,
    author       = {Zhengyao Jiang and Dixing Xu and Jinjun Liang},
    title        = {A Deep Reinforcement Learning Framework for the Financial Portfolio Management Problem},
    year         = {2017},
    eprint       = {1706.10059},
    archivePrefix = {arXiv},
    primaryClass = {q-fin.CP},
}

@inproceedings{Lillicrap2016continuous,
    author    = {Timothy P. Lillicrap and Jonathan J. Hunt and Alexander Pritzel and Nicolas Heess and Tom Erez and Yuval Tassa and David Silver and Daan Wierstra},
    title     = {Continuous Control with Deep Reinforcement Learning},
    booktitle = {Proceedings of the 4th International Conference on Learning Representations ({ICLR} 2016)},
    year      = {2016},
    address   = {San Juan, Puerto Rico},
}

@inproceedings{Liu2018DRLPortOpt,
    author    = {Xiao-Yang Liu and Zhuoran Xiong and Shan Zhong and Hongyang Yang and Anwar Walid},
    title     = {Practical Deep Reinforcement Learning Approach for Stock Trading},
    booktitle = {Proceedings of the NeurIPS 2018 Workshop on Challenges and Opportunities for {AI} in Financial Services},
    year      = {2018},
    month     = dec,
    address   = {Montr\'{e}al, Canada},
}

@article{Lo2000Foundations,
    author  = {Andrew W. Lo and Harry Mamaysky and Jiang Wang},
    title   = {Foundations of Technical Analysis: Computational Algorithms, Statistical Inference, and Empirical Implementation},
    journal = {Journal of Finance},
    year    = {2000},
    volume  = {55},
    number  = {4},
    pages   = {1705--1765},
}

@misc{LopezLira2025ChatGPT_Forecast,
    author       = {Alejandro Lopez-Lira and Yuehua Tang},
    title        = {Can {ChatGPT} Forecast Stock Price Movements? {R}eturn Predictability and Large Language Models},
    year         = {2025},
    eprint       = {2304.07619},
    archivePrefix = {arXiv},
    primaryClass = {q-fin.ST},
}

@article{MnihDQN,
    author  = {Volodymyr Mnih and Koray Kavukcuoglu and David Silver and Andrei A. Rusu and Joel Veness and Marc G. Bellemare and Alex Graves and Martin Riedmiller and Andreas K. Fidjeland and Georg Ostrovski and Stig Petersen and Charles Beattie and Amir Sadik and Ioannis Antonoglou and Helen King and Dharshan Kumaran and Daan Wierstra and Shane Legg and Demis Hassabis},
    title   = {Human-Level Control through Deep Reinforcement Learning},
    journal = {Nature},
    year    = {2015},
    volume  = {518},
    pages   = {484--489},
}

@inproceedings{pmlr-v48-mniha16,
    author    = {Volodymyr Mnih and Adria Puigdomenech Badia and Mehdi Mirza and Alex Graves and Timothy Lillicrap and Tim Harley and David Silver and Koray Kavukcuoglu},
    title     = {Asynchronous Methods for Deep Reinforcement Learning},
    booktitle = {Proceedings of the 33rd International Conference on Machine Learning ({ICML})},
    year      = {2016},
    volume    = {48},
    series    = {Proceedings of Machine Learning Research},
    pages     = {1928--1937},
    publisher = {PMLR},
    address   = {New York, NY, USA},
}

@article{Moody1998DRL_Trading,
    author  = {John Moody and Lizhong Wu and Yuansong Liao and Matthew Saffell},
    title   = {Performance Functions and Reinforcement Learning for Trading Systems and Portfolios},
    journal = {Journal of Forecasting},
    year    = {1998},
    volume  = {17},
    number  = {5--6},
    pages   = {441--470},
}

@book{Murphy1999Technical,
    author    = {John J. Murphy},
    title     = {Technical Analysis of the Financial Markets},
    publisher = {New York Institute of Finance},
    year      = {1999},
}

@inproceedings{Ng1999PolicyInvariance,
    author    = {Andrew Y. Ng and Daishi Harada and Stuart J. Russell},
    title     = {Policy Invariance Under Reward Transformations: Theory and Application to Reward Shaping},
    booktitle = {Proceedings of the Sixteenth International Conference on Machine Learning ({ICML})},
    year      = {1999},
    pages     = {278--287},
    publisher = {Morgan Kaufmann},
}

@misc{qian2025whenagentstrade,
  title = {When Agents Trade: Live Multi-Market Trading Benchmark for {LLM} Agents},
  author = {Lingfei Qian and Xueqing Peng and Yan Wang and Vincent Jim Zhang and Huan He and Hanley Smith and Yi Han and Yueru He and Haohang Li and Yupeng Cao and Yangyang Yu and Alejandro Lopez-Lira and Peng Lu and Jian-Yun Nie and Guojun Xiong and Jimin Huang and Sophia Ananiadou},
  year = {2025},
  eprint = {2510.11695},
  archivePrefix = {arXiv},
  primaryClass = {cs.CL},
}

@misc{Schulman2017PPO,
    author       = {John Schulman and Filip Wolski and Prafulla Dhariwal and Alec Radford and Oleg Klimov},
    title        = {Proximal Policy Optimization Algorithms},
    year         = {2017},
    eprint       = {1707.06347},
    archivePrefix = {arXiv},
    primaryClass = {cs.LG},
}

@book{Sutton1998,
    author    = {Richard S. Sutton and Andrew G. Barto},
    title     = {Reinforcement Learning: An Introduction},
    edition   = {Second},
    publisher = {The MIT Press},
    year      = {2018},
}

@article{dubey2024llama3,
  title   = {The {LLaMA} 3 Herd of Models},
  author  = {Aaron Grattafiori and Abhimanyu Dubey and Abhinav Jauhri and Abhinav Pandey and others},
  year={2024},
  eprint={2407.21783},
  archivePrefix={arXiv},
  primaryClass={cs.AI},
}

@inproceedings{vanHasselt2015DoubleDQL,
    author    = {Hado van Hasselt and Arthur Guez and David Silver},
    title     = {Deep Reinforcement Learning with Double {Q}-Learning},
    booktitle = {Proceedings of the Thirtieth AAAI Conference on Artificial Intelligence ({AAAI-16})},
    year      = {2016},
    publisher = {Association for the Advancement of Artificial Intelligence},
    pages = {2094–2100},
    numpages = {7},
    address = {Phoenix, Arizona},
}

@book{Wilder1978NewTechnical,
    author    = {J. Welles Wilder},
    title     = {New Concepts in Technical Trading Systems},
    publisher = {Trend Research},
    year      = {1978},
}

@article{Williams1992REINFORCE,
    author  = {Ronald J. Williams},
    title   = {Simple Statistical Gradient-Following Algorithms for Connectionist Reinforcement Learning},
    journal = {Machine Learning},
    year    = {1992},
    volume  = {8},
    number  = {3--4},
    pages   = {229--256},
}

@inproceedings{Xie2026clef2026finmmeval,
  title = {Overview of {FinMMEval} 2026: Multilingual and Multimodal Financial Evaluation},
  author = {Zhuohan Xie and Yuyang Dai and Rania Elbadry and Vanshikaa Jani and Xueqing Peng and Lingfei Qian and Georgi Georgiev and Dimitar Dimitrov and Fan Zhang and Jimin Huang and Jiahui Geng and Yankai Chen and Ye Yuan and Haolun Wu and Yuxia Wang and Ivan Koychev and Veselin Stoyanov and Mingzi Song and Yu Chen and Steve Liu and Preslav Nakov},
  booktitle = {Experimental IR Meets Multilinguality, Multimodality, and Interaction},
  series = {Proceedings of the Seventeenth International Conference of the CLEF Association (CLEF 2026)},
  year = {2026},
  month = {September 21--24},
  address = {Jena, Germany},
  publisher = {Springer Lecture Notes in Computer Science LNCS},
}

@inproceedings{Yang2020EnsembleTrading,
    author    = {Hongyang Yang and Xiaoyang Liu and Shan Zhong and Anwar Walid},
    title     = {Deep Reinforcement Learning for Automated Stock Trading: An Ensemble Strategy},
    booktitle = {Proceedings of the ACM International Conference on AI in Finance ({ICAIF})},
    year      = {2020},
    month     = oct,
    address   = {New York, NY, USA},
}

@misc{Ye2024SentimentBasedEnsemble,
    author       = {Andrew Ye and James Xu and Vidyut Veedgav and Yi Wang and Yifan Yu and Daniel Yan and Ryan Chen and Vipin Chaudhary and Shuai Xu},
    title        = {Learning the Market: Sentiment-Based Ensemble Trading Agents},
    year         = {2024},
    eprint       = {2402.01441},
    archivePrefix = {arXiv},
    primaryClass = {q-fin.TR},
}

@inproceedings{Yu2023LLM_Prediction,
    author    = {Xinli Yu and Zheng Chen and Yanbin Lu},
    title     = {Harnessing {LLM}s for Temporal Data -- A Study on Explainable Financial Time Series Forecasting},
    booktitle = {Proceedings of the 2023 Conference on Empirical Methods in Natural Language Processing ({EMNLP}): Industry Track},
    year      = {2023},
    month     = dec,
    address   = {Singapore},
    pages     = {739--753},
}

\appendix



\end{document}